\documentclass[letterpaper, 10 pt, conference]{ieeeconf}  

\IEEEoverridecommandlockouts                              

\usepackage{graphics} 
\usepackage{amsmath} 
\usepackage{graphicx}
\usepackage{subfig}
\usepackage{hyperref}
\usepackage{graphicx}
\usepackage{caption}

\title{\LARGE \bf
Combining Self-Supervised Learning and Imitation for Vision-Based Rope Manipulation
}

\author{Ashvin Nair$^{*}$ \hspace{5mm} Dian Chen$^{*}$ \hspace{5mm}  Pulkit Agrawal$^{*}$\thanks{$^{*}$Equal contribution.}\\ 
Phillip Isola \hspace{5mm} Pieter Abbeel \hspace{5mm} Jitendra Malik \hspace{5mm} Sergey Levine
\thanks{The authors are with the Department of Electrical Engineering and Computer Science, University of California, Berkeley, CA, USA.%
}
}

\begin{document}

\maketitle
\thispagestyle{empty}
\pagestyle{empty}

\begin{abstract}
Manipulation of deformable objects, such as ropes and cloth, is an important but challenging problem in robotics.
We present a learning-based system where a robot takes as input a sequence of images of a human manipulating a rope from an initial to goal configuration, and outputs a sequence of actions that can reproduce the human demonstration, using only monocular images as input. To perform this task, the robot learns a pixel-level inverse dynamics model of rope manipulation directly from images in a self-supervised manner, using about 60K interactions with the rope collected autonomously by the robot. The human demonstration provides a high-level plan of what to do and the low-level inverse model is used to execute the plan. We show that by combining the high and low-level plans, the robot can successfully manipulate a rope into a variety of target shapes using only a sequence of human-provided images for direction.

\end{abstract}

\section{Introduction}

Manipulation of deformable objects, such as ropes and cloth, is an important but challenging problem in robotics. Open-loop strategies for deformable object manipulation are often ineffective, since the material can shift in unpredictable ways \cite{hopcroft1991case}. Perception of cloth and rope also poses a major challenge, since standard methods for estimating the pose of rigid objects cannot be readily applied to deformable objects for which it is difficult to concretely define the degrees of freedom or provide suitable training data \cite{miller2011parametrized}. Despite the numerous industrial and commercial applications that an effective system for deformable object manipulation would have, effective and reliable methods for such tasks remain exceptionally difficult to construct. Previous work on deformable object manipulation has sought to use sophisticated finite element models \cite{hopcroft1991case,bell2010flexible}, hand-engineered representations \cite{kuniyoshi1994learning,wakamatsu2006knotting,morita2003knot,maitin2010cloth}, and direct imitation of human-provided demonstrations \cite{mayer2008system,schulman2013warping}. Direct model identification for ropes and cloth is challenging and brittle, while imitation of human demonstrations without an internal model of the object's dynamics is liable to fail in conditions that deviate from those in the demonstrations.

\begin{figure*}[t]%
    \centering
    \includegraphics[width=0.9\linewidth]{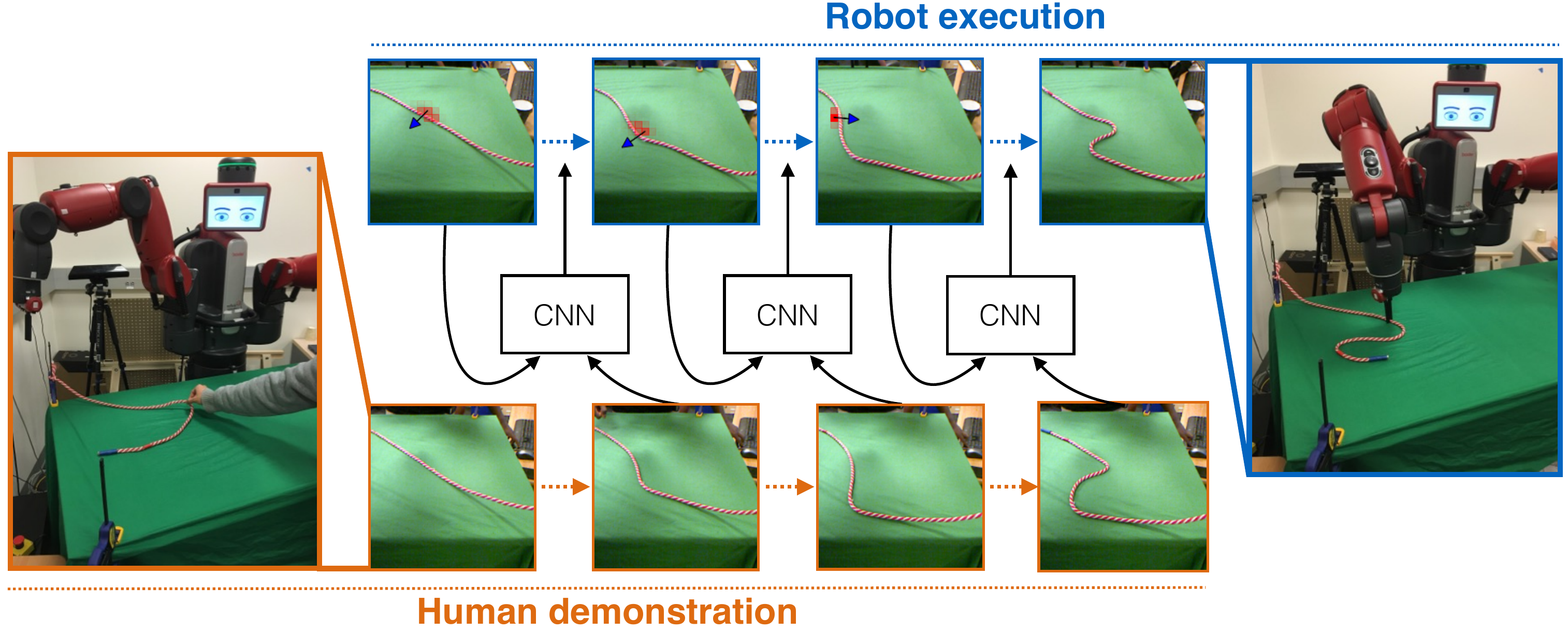}
    \caption{We present a system where the robot is capable of manipulating a rope into target configurations by combining a high-level plan provided by a human with a learned low-level model of rope manipulation. A human provides the robot with a sequence of images recorded while he manipulates the rope from an initial to goal configuration. The robot uses a learned inverse dynamics model to execute actions to follow the demonstrated trajectory. The robot uses a convolutional neural network (CNN) for learning the inverse model in a self-supervised manner using 60K interactions with the rope with no human supervision. The red heatmap on each image of the robot's execution trace shows the predicted location of the \textit{pick} action and the blue arrow shows the direction of the action. This image is best seen in color.}%
    \label{fig:teaser}%
\end{figure*}

In this work, we instead propose a learning-based approach to associate the behavior of a deformable object with a robot's actions, using self-supervision from large amounts of data gathered autonomously by the robot. In particular, the robot learns a goal-directed inverse dynamics model: given a current state and a goal state (both in image space), it predicts the action that will achieve the goal. Once this model is learned, our method can use human-provided demonstrations as higher level guidance. In effect, the demonstrations tell the robot \emph{what} to do, while the learned model tells it \emph{how} to do it, combining high-level human direction with a learned model of low-level dynamics. Figure \ref{fig:teaser} shows an overview of our system. 

Our method does not use any explicit parameterization of the rope configuration. Instead, we learn a model using raw images of the rope, which provides representational flexibility and avoids the need for manually specifying kinematic models that may fail to adequately capture the full variability of the deformable object. To handle high-dimensional visual observations, we employ deep convolutional neural networks for learning the inverse dynamics model.

Many interesting manipulation tasks require more than a single step to achieve a desired goal state. For example, tying a rope into a knot, stitching tissue during a surgery, and lacing shoes all involve multiple steps of manipulation.

Learning to perform these tasks is much harder than learning to perform small deformations, because only a very specific set of chained actions will result in success. While self-supervised learning of chained actions remains an open challenge, here we alleviate the problem by employing a small amount of imitation. At test time, the robot is provided a sequence of images depicting each step in the task. Because it has learned how to achieve small deformations between current and target images, it can follow the sequence and complete a multistep task.

One of the key challenges in self-supervised robot learning is collecting enough data for learning skilled behaviors, since the state and action space for practical manipulation tasks is extremely large. In comparison to rigid objects, deformable objects such as ropes can take on a much larger range of configurations. This means that for self-supervised learning of rope manipulation we require large amounts of interaction data. To this end, we configured a Baxter robot to autonomously collect interaction data with a rope without any human intervention. To date, we have collected over 500 hours worth of data which has been publicly released at the \href{https://ropemanipulation.github.io/}{\texttt{project website}}.

The main contribution of our work is to present a learning-based approach for rope manipulation that combines learned predictive models with high-level human-provided demonstrations. The predictive model is learned autonomously by the robot, using automatically collected data and self-supervision. This model allows the robot to understand \emph{how} to manipulate a rope to reach target configurations. The human demonstrations, in the form of step-by-step images of a rope undergoing a manipulation task, can then be used at test time to indicate to the robot \emph{what} should be done to the rope, and the learned model can then be used to determine \emph{how} to do it. We evaluate our method on a Baxter robot trained on our dataset of over 500 hours of real-world rope manipulation, and demonstrate the ability of the system to arrange a rope into a variety of shapes using high-level human direction.

\section{Related Work}
There has been significant recent interest in learning control policies directly from visual inputs using deep neural networks. Impressive results have been obtained on playing Atari games \cite{mnih2015human}, a large suite of robotic tasks in simulation \cite{lillicrap2015continuous,mnih2016asynchronous,billiards,wahlstrom2015from,watter2015embed,actionconditioned} and real world robotic manipulation tasks involving rigid objects in controlled settings \cite{LevineFDA15,lenz2015deepMPC}.
Given that state of the art deep reinforcement learning algorithms are data hungry, some recents works learned to grasp \cite{pinto2015supersizing,levine2016learning} and push \cite{agrawal2016poking} real world objects by collecting large amounts of robot interaction data. 

Manipulating deformable objects has been of great interest to the robotics community \cite{hopcroft1991case}. Prior works have considered problems such as surgical suturing \cite{mayer2008system,schulman2013warping}, towel folding \cite{maitin2010cloth},
knot tying and rope manipulation among many others. Rope manipulation and knot tying are most closely related to our work. Inoue et al. \cite{inoue1985hand} investigated the problem of knot tying and following works used motion planning \cite{saha2006motion}, fixtures for open-loop execution \cite{bell2010flexible} and robotic hands with tactile feedback \cite{yamakawa2007one}. Morita et al. \cite{morita2003knot} developed a system for tying knots from visual observations that makes use of knot theory \cite{crowell2012introduction} to identify a sequence of knot states and then execute motor primitives to achieve these states. Wakamatsu et al. \cite{wakamatsu2006knotting} chose a sequence of robotic hand motion primitives from rope cross states inferred by a planner to achieve a desired knot from a given initial state of the rope.  In contrast to these works, our goal is not to tie knots but to manipulate rope into a general configuration by watching a human as she manipulates the rope. Our system does not require any rope-specific knowledge and is therefore generalizable to  manipulating other deformable objects.  

Schulman et al. \cite{schulman2013generalization} used non-rigid registration \cite{chui2000tpsrpm} for transferring human demonstrated trajectories for rope manipulation. In the learning from demonstration (LFD) paradigm, an expert provides the robot with examples of observations and the associated motor trajectories used to perform the task. The robot then builds a model from this data that outputs a policy for achieving the same task at test time. A survey of learning from demonstration can be found in \cite{argall2009survey}. One drawback of the LFD approach is that, if a robot is to perform a wide suite of tasks, an expert is required to provide demonstrations for each task individually. Instead, if robots learn to imitate human behavior by simply observing humans as they perform different tasks, robots could learn much faster. Many past works have proposed solutions to learning from visual demonstrations \cite{lee2013syntactic,yang2015robot,kuniyoshi1994learning}.

The closest to our work are Yang et al. \cite{yang2015robot} and Kuniyoshi et al. \cite{kuniyoshi1994learning}. Kuniyoshi et al. use a vision system to detect an object and the exact pose of human hands to produce a sequence of robotic actions to copy the human demonstration. Yang et al. predict object detections and grasp types that are then used to infer the robotic actions using an action parse tree. In contrast to these approaches that deal only with rigid objects, we consider the significantly harder problem of manipulating a deformable object. In addition to integrating human demonstrations, we let the robot collect supervised data to build a predictive model of its actions, which can then be used to infer how the robot can imitate the visual demonstration provided by the expert.

\section{Experimental Setup}

\begin{figure}%
    \centering
    \includegraphics[width=1.0\linewidth]{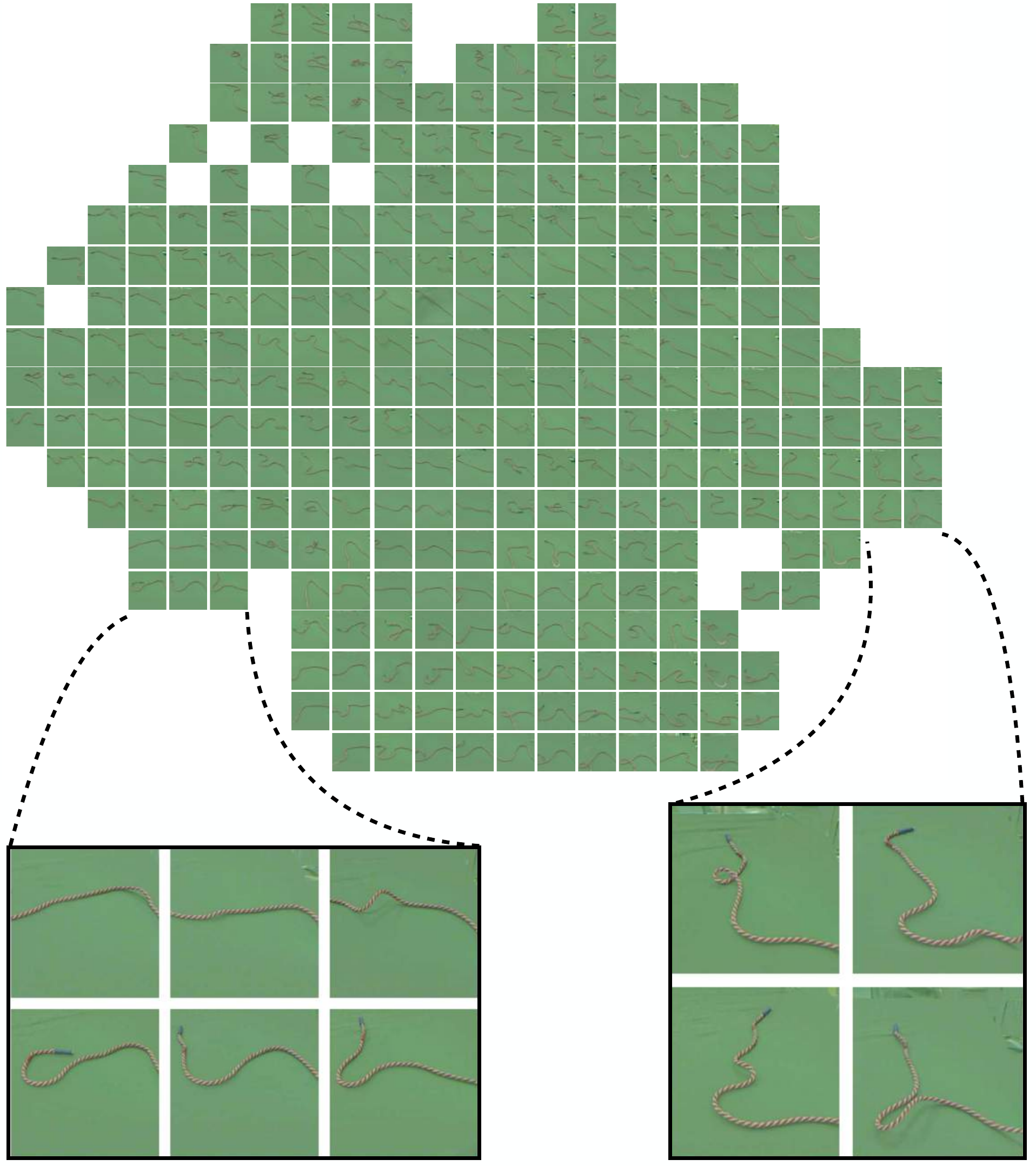}
    \caption{A sample of validation set images visualized using t-SNE \cite{maaten2008visualizing} over the image features learned by our model. Notice that similar configurations of the rope appear near each other, indicating the learned feature space meaningfully organizes variation in rope shape.}%
    \label{fig:dataset}%
\end{figure}

We use a Baxter robot for all experiments described in the paper. The robot manipulates a rope placed on a table in front of it using only one arm.
The arm has a parallel jaw gripper that can rotate, open and close. One end of the rope is tied to a clamp attached to the table. The robot receives visual inputs from the RGB channels of a Kinect camera. The setup is illustrated in Figure~\ref{fig:teaser}. The interaction of the robot with the rope is constrained to a single action primitive consisting of two sub-actions - \textit{pick} the rope at location $(x_1,y_1)$ and \textit{drop} the the rope at location $(x_2, y_2)$, where $(x_1, y_1, x_2, y_2)$ are pixel coordinates in the input RGB image. It is possible to manipulate the rope into many complex configurations using just this action primitive, as show in Figure~\ref{fig:dataset}.

The robot collects data in a self-supervised manner by randomly choosing pairs of \textit{pick} and \textit{drop} points in the image. However, if we randomly choose a point on the image, then most points will not be on the rope. Instead, we use the point cloud from the Kinect to segment the rope and then choose a \textit{pick} point uniformly at random from this segment.
Once the \textit{pick} point is chosen, the \textit{drop} point can be obtained as a displacement vector from the \textit{pick} point. We represent this displacement vector by the angle $\theta \in [0, 2\pi)$ and length $l \in [1, 15]$ cm. Values of $\theta$ and $l$ are uniformly and randomly sampled from their respective ranges to obtain the \textit{drop} point.
After choosing the pick and drop points, the robot executes the following steps: (1) grasp the rope at the \textit{pick} point, (2) move the arm 5 cm vertically above the pick point, (3) move the arm to a point 5 cm vertically above the \textit{drop} point, (4) release the rope by opening the gripper. The pair of current and next image obtained after executing the random action are used for training the inverse model described in Section \ref{sec:method}. 

During autonomous data collection, it is very likely that the rope will fall off the table or drift out of reach of the robot and consequently halt the data collection process until a manual reset if performed. For continuous collection of data without human intervention, the robot performs a reset action either after every 50 actions or if there are fewer than 1000 pixels of rope in the robot's workspace.
The reset action detects the end of the rope and then pulls the rope in a fixed direction to make it straight. This system can run continuously for stretches of more than twenty hours without any manual intervention. 

\section{Method for Rope Manipulation}
\label{sec:method}

\begin{figure*}[t]%
    \centering
    \includegraphics[width=0.9\linewidth]{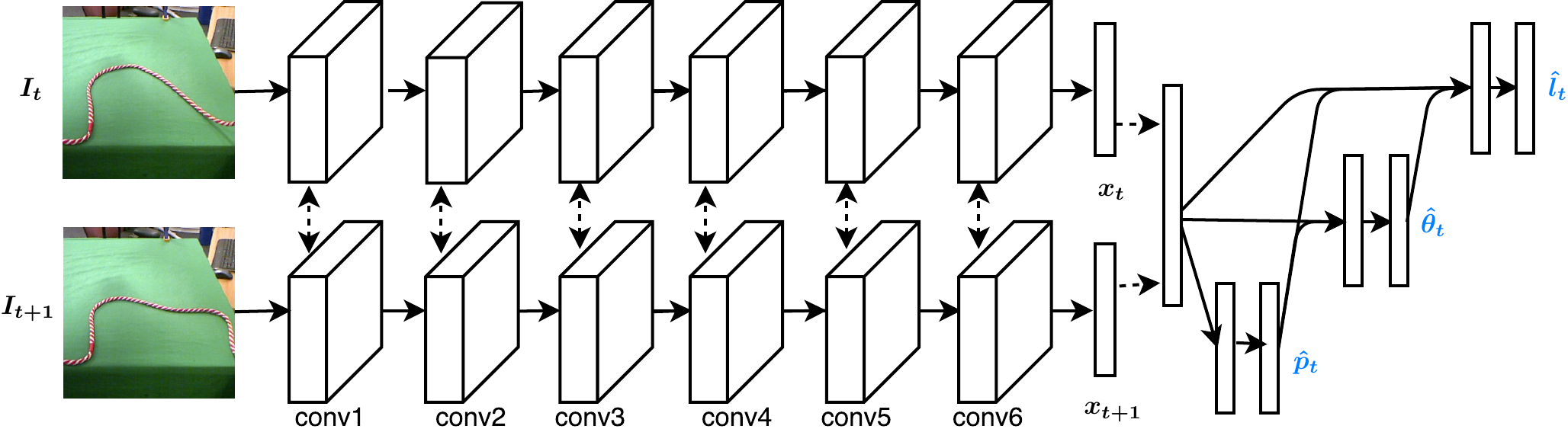}
    \caption{We use a convolutional neural network (CNN) to build the inverse dynamics model. The input to the CNN is a pair of images $(I_t, I_{t+1})$ and the output is the action that moves the rope from the shape in $I_t$ into the shape in $I_{t+1}$. The action is parameterized as $(p_t, \theta_t, l_t)$, where  $p_t, \theta_t, l_t$ is the action location, direction, and length, respectively. $\hat{p}_t, \hat{\theta_t}, \hat{l}_t$ denote the predictions. The CNN consists of two streams with shared weights that transform each image into a latent feature space $x$. The representations $x_t, x_{t+1}$ are concatenated together and fed into fully-connected layers to predict the action. We use approximately 60K rope interactions for training the CNN.}%
    \label{fig:network}%
\end{figure*}

Our goal is to have the robot watch a human manipulate a rope and then reproduce this manipulation on its own. The human provides a demonstration in the form of a sequence of images of the rope in intermediate states toward a final goal state. Let V = $\{I_t | t = 1..T\}$ represent this sequence. The robot must execute a series of actions to transform $I_1$ into $I_2$, then $I_{2}$ into $I_{3}$, and so on until the end of the sequence.

A model that predicts the action that relates a pair of input states is called an inverse dynamics model, and is mathematically described in equation \ref{eq:inverse} below:
\begin{equation}
    \label{eq:inverse}
    {u}_{t} = F(I_t, I_{t+1}),
\end{equation}
where $I_t$ and $I_{t+1}$ are images of the current and next states and $u_t$ is the action. 
We use convolutional neural networks to learn the inverse model in a manner similar to \cite{agrawal2016poking,pinto2016curious}. Details of training the inverse model and the process of imitating an observed visual demonstration are described in Sections \ref{sec:inverse} and \ref{sec:imitation}, respectively. 

\subsection{Neural Network Architecture Notation} 

Let the abbreviations Ck, Fk represent a convolutional (C) layer with k filters, a fully-connected (F) layer with k filters respectively. We used the ELU (Exponential Linear unit) non-linearity after every convolutional/fully-connected layer, except for the output layer. As an example of our notation, C48-F500 refers to a network with 48 filters in the convolution layer followed by ELU non-linearity and a fully-connected layer with 500 units.

\subsection{Self-Supervised Learning of the Inverse Model}

\label{sec:inverse}
Our neural network architecture consists of two streams that transform each of the two input images into a latent feature space, $x$. The architecture of these  streams is C96-C256-C384-C384-C256-C200. The first five out of the six layers have the same architecture as AlexNet. The neural network weights of both the streams are tied. The latent representations of the two images, $(x_t, x_{t+1})$, each size 200, are concatenated and fed into another neural network with the architecture F200-F200. This subnetwork provides a joint non-linear feature representation of the current and next image that is used to predict the action.

For the purpose of training, we turn action prediction into a classification problem by discretizing the action space. The action is parameterized as a tuple $(p_t, \theta_t, l_t)$, where $p_t$ is the action location, $\theta_t$ is the action direction and $l_t$ is the action length. Each dimension of this tuple is independently discretized. The action location is discretized onto a $20\times20$ spatial grid, and the direction and length are discretized into 36 and 10 bins respectively. The predicted action tuple is represented as $(\hat{p}_t, \hat{\theta}_t, \hat{l}_t)$ Treating action prediction as classification makes training the network easier and accounts for multimodality in the output distribution, which can occur when multiple actions exist that move the rope from the same initial to final configuration.

A na\"{i}ve way of training the neural network would be to classify each of the three action elements independently. In order to model the joint discrete distribution of the action $P(p_t, \theta_t, l_t)$ in a way that does not increases exponentially in size with the number of dimensions, we decompose the joint distribution as $P(p_t, \theta_t, l_t) = P(p_t)P(\theta_t|p_t)P(l_t|\theta_t, p_t)$. The neural network first predicts a distribution over 400 different possible pick locations. The $argmax$ of this distribution ($\hat{p}_t$) is chosen as the pick location. One-hot encoding of the pick location is passed along with state features ($x_t, x_{t+1}$) to predict a distribution over possible action directions $(\theta_t)$. Then the $argmax$ over the action direction distribution ($\hat{\theta}_t$) is one-hot encoded and concatenated with the one-hot encoding of $\hat{p}_t$ and state features to predict $\hat{l}_t$.

We initialize the first five layers of the network using pre-trained AlexNet weights obtained by training for image classification on the ImageNet challenge \cite{krizhevsky2012imagenet}. For the first 5K iterations we set the learning rate of these convolutional layers to be 0. For the rest of the training we set the learning rate to 1e-4, and use the Adam optimizer \cite{kingma2014adam}. All the other layers are initialized with small normally distributed weights and trained from the first iteration with a learning rate of 1e-4. Approximately 60K pairs of before and after image collected autonomously by the robot were used for training the inverse model. A separate validation set of 2.5K before/after image pairs was used for hyper-parameter tuning. 

\subsection{Imitating Human Demonstration}
\label{sec:imitation}

\begin{figure*}[t]
    \centering
    \includegraphics[width=1.0\linewidth]{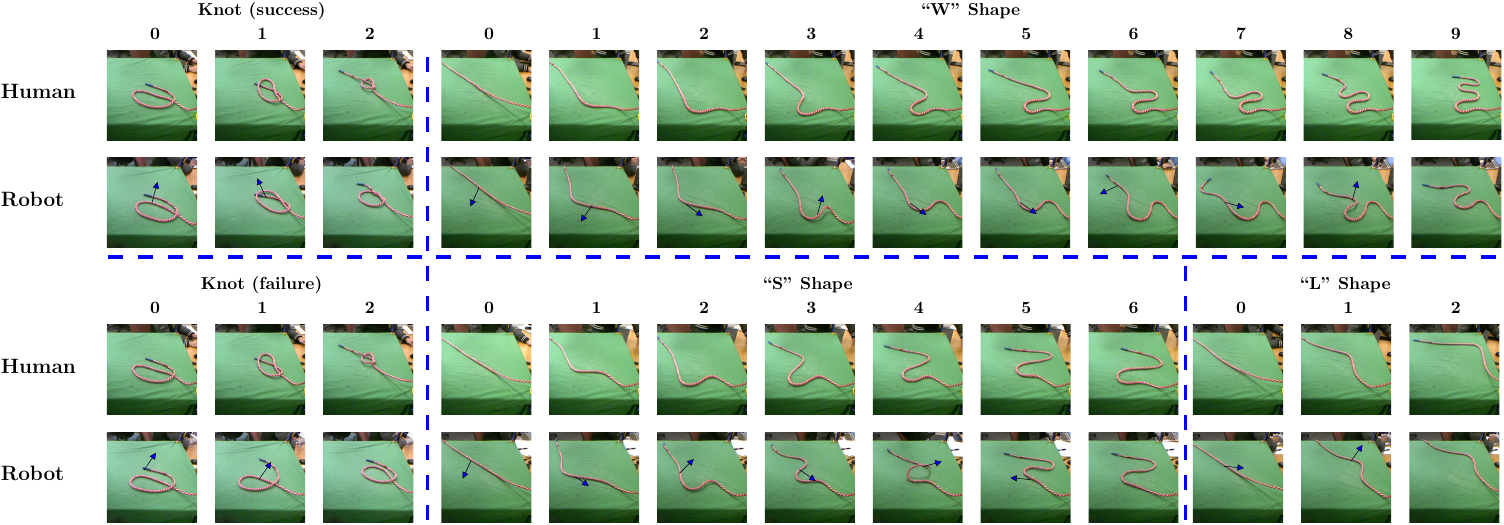}
    \caption{Qualitative comparison of the robot's performance in imitating the human demonstration for arranging the rope into \textit{W, S, L} and \textit{knot} configurations. The upper row in each example shows the sequence of demonstration images provided as input to the robot and the second (lower) row shows the states achieved by the robot as it tries to follow the demonstrated trajectory. The blue arrow on each image of the robot's execution trace shows the direction and the location of the pick point of the action performed by the robot. Please see the supplementary materials on the \href{https://ropemanipulation.github.io/}{project website} for more examples.}%
    \label{fig:results}
\end{figure*}

With the help of an inverse model, the robot can deform the rope by small amounts. Different specific sequences of such small deformations can be used to manipulate the rope in different ways. However, learning a chain of actions that manipulate the rope into configurations of interest such as knots, is non-trivial. In this work we mitigate this challenge by using visual demonstrations from a human. The robot receives as input the sequence of images depicting each stage of the manipulation performed by a human demonstrator to reach a desired rope configuration from an initial rope configuration. We denote this sequence of demonstration images as $V = \{I'_t | t \in (1\dots T)\} $ where $I'_T$ depicts the goal. 

Let $I_1$ be the initial image the $I'{_t}$ be the goal image. The robot first inputs the pair of images $(I_1, I'_2)$ into the learned inverse model and executes the predicted action.
Let $I_2$ be the visual state of the world after the action is executed. The robot then inputs $(I_2, I'_3)$ into the inverse model and executes the output action. This process is repeated iteratively for $T$ time steps. In some cases the robot predicts a pick location that does not lie on the rope. For these cases we use the rope segmentation information to find the point on the rope that is closest to predicted pick location to execute the pick primitive.

\subsection{Active Data Collection}
\label{sec:scaling}
With randomized data collection it is unlikely that the robot would place the rope in interesting configurations and therefore the model may not be accurate at performing tasks such as knot-tying that require moving the rope into complex configurations. In order to bias data collection towards interesting configurations, we collected a set of 50 images of the rope when it was manually arranged in a random configuration (i.e. the goal buffer). We then used a model trained with 30K randomly collected data points and instead of randomly sampling an action, we randomly sampled a image from the goal buffer and set that as the goal image. We passed the current and goal image into the inverse model and used the action predicted by the inverse model for data collection. 

\section{Evaluation Procedure}

The performance of the robot was evaluated by tasking it to reconfigure the rope from a given initial configuration into target configurations of varying complexity depicting ``L", ``S", ``W" shapes and a knot. 

The performance was quantitatively measured by calculating the distance between the rope configurations in the sequence of images provided from the human demonstration and the sequence of images achieved by the robot after executing the actions from the inverse dynamics model. The distance between two rope configurations is computed by first segmenting the rope in each image, then aligning points in these two segmentations with thin plate spline robust point matching (TPS-RPM) \cite{chui2000tpsrpm} and calculating the mean pixel distance between the matched points. We compare the performance of our method against a hand-engineered baseline, a nearest neighbor baseline, and our method without imitation (see Section \ref{sec:baseline}). Videos of the self-supervised data collection, the demonstrations, and autonomous executions are available at \href{https://ropemanipulation.github.io/}{\texttt{https://ropemanipulation.github.io/}}

\subsection{Baseline}
\label{sec:baseline}
\noindent \textbf{Hand-Engineered baseline:} The first baseline we compare against is a hand-engineered method that takes as input the sequence of images from the human demonstration. For inferring which action should be executed to transform the rope from the configuration in $I_t$ into the configuration in $I_{t+1}$, we first segment the rope in both the images, and use TPS-RPM to register the segments. In the absence of a model of rope dynamics, a simple way to move the rope into a target configuration is to \textit{pick} the rope at the point with the largest deformation in the first image relative to the second and then \textit{drop} the rope at the corresponding point in the second image. As the point with largest distance may be an outlier, we use the point at the $90^{th}$ percentile of the deformation distances for the \textit{pick} action.

\noindent \textbf{Nearest Neighbor baseline:} To evaluate whether the neural network simply memorized the training data, we compared our method to a nearest neighbor baseline. Given the current image ($I_t$) and the target image in the human demonstration ($I'_{t+1}$), a pair of images $(I_k, I_{k+1})$ in the training set that is closest to $(I_t, I'_{t+1})$ is determined and the ground truth action used to collect this training sample is executed by the robot. As the distance metric for nearest neighbor calculation, we used Euclidean distance in raw RGB space after down-sampling images to $32\times32$. 

\begin{figure}%
    \centering
    \includegraphics[width=8cm]{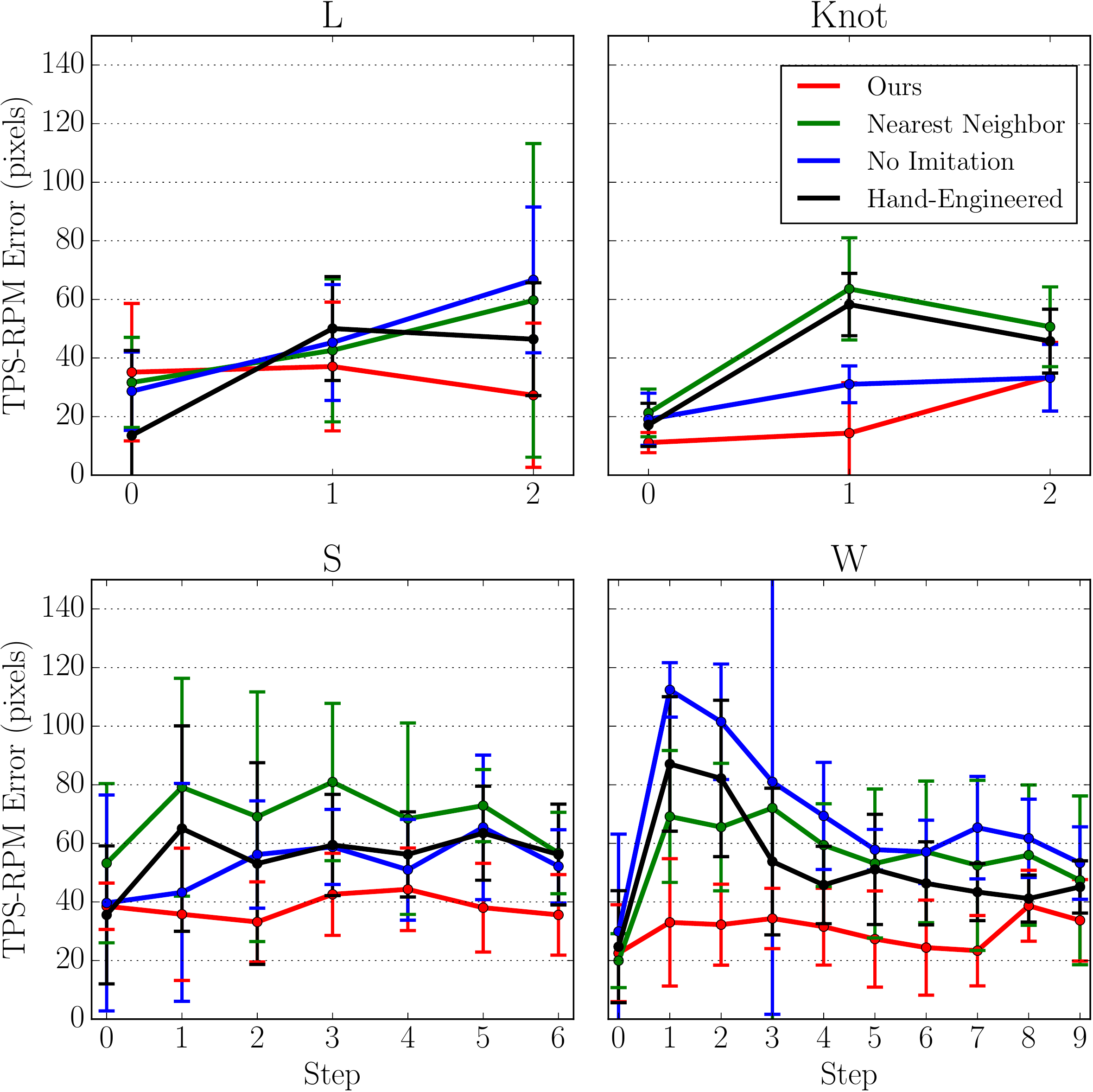}
    \caption{Performance comparison of the proposed method against the baselines described in Section \ref{sec:baseline}. Performance is measured as the TPS-RPM distance metric between the rope configurations in the sequence of images achieved by the robot and the ones provided by human demonstration. Lower distance indicates better performance. Each subplot shows performance for a different target shape. Our method outperforms the baseline methods.}%
    \label{fig:trajerror}%
\end{figure}

\noindent \textbf{No Imitation baseline:} For understanding how critical is imitation for manipulating the rope into desired shape, we evaluated the learned model by feeding in only the initial and goal image ($I_1,I_T'$) without any intermediate steps. The inverse model takes these images as inputs and the robot executes the predicted action and image $I_2$ is obtained. Next, the pair ($I_2,I_T'$) is fed into the inverse model to infer the next action to execute and this process is repeated iteratively for the same number of steps as in the human demonstration.

\begin{figure}
    \centering
    \includegraphics[width=\linewidth]{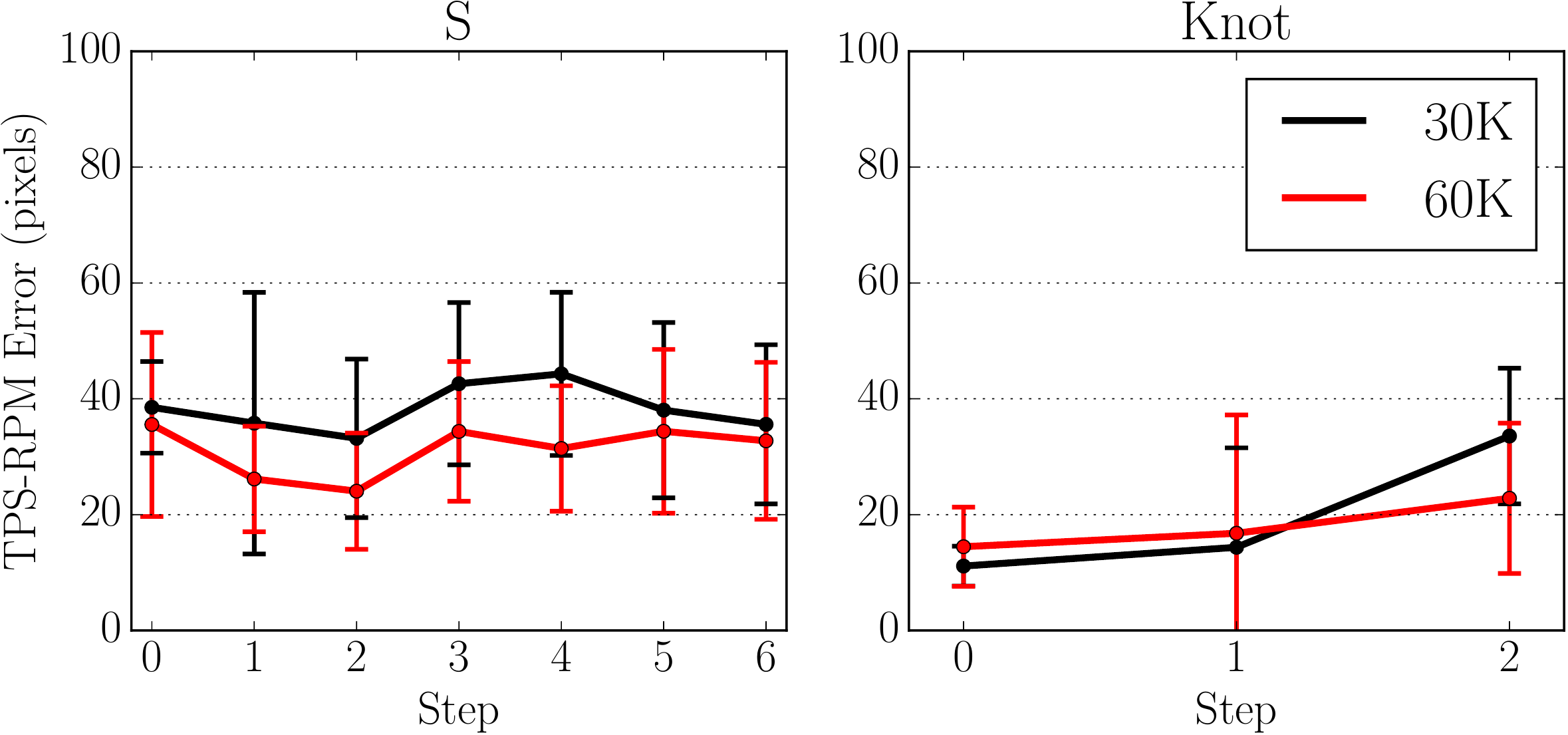}
    \\
    Knot Tying Success Rate
    
    \begin{tabular}{ | c || c | c |}
        \hline
        Imitation? & 30K & 60K  \\ \hline
        Yes & 6/50 & 19/50  \\ \hline
        No  & 5/50 & 11/50  \\ \hline
    \end{tabular}
    
    \caption{The performance of our method improves both in manipulating the rope in desired shapes and tying knots. The results indicate that with larger amounts of data even better performance can be achieved.}
    \label{fig:scaling}%
\end{figure}

\section{Results}
Figure \ref{fig:results} qualitatively shows that using the learned inverse dynamics model, the robot is capable of re-configuring the rope into many different shapes. It can also be seen that when the rope is not bent sharply, the robot is more accurate at manipulation. 

Figure \ref{fig:trajerror} compares the performance of our method against the hand-engineered, nearest neighbor and no-imitation baselines described in Section \ref{sec:baseline}. Each subfigure shows the performance for a different target configuration.
The x-axis in each subfigure corresponds to the number of intermediate images that were provided to the robot via demonstration. The y-axis corresponds to the TPS-RPM distance between the rope configuration achieved by the robot and the corresponding rope configuration in the actual demonstration. Lower values indicate better performance. 

For every sequence, mean accuracy and standard deviation are reported across 10 different repeats of two human demonstration sequences. The results demonstrate that our method outperforms various baselines including a hand-engineered baseline (i.e. TPS-RPM baseline), indicating that through self-supervision, the robot has learned a dynamics model of rope motion that is useful for manipulation.

\subsection{Scaling with amount of data}
Figure \ref{fig:scaling} shows that a model trained with 60K data points out of which 30K were collected using random sampling and other 30K using active sampling significantly outperforms a model trained with only 30K data points. Our method can successfully tie knots 38\% of the time. Due to the lack of 60K randomly collected training points, at the moment we cannot conclude how useful is active data sampling as compared to random sampling. We will include these results in the next revision of our paper. 

\subsection{Importance of Imitation} 
How important are human demonstrations for manipulating the rope into a desired configuration? Results in Figure ~\ref{fig:trajerror} show that when the robot was only provided the initial and final images instead of all the images in the human demonstration sequence, the performance is significantly worse. Figure \ref{fig:scaling} further shows that without imitation robot is able to tie the knot only 11/50 times instead of 19/50 times with imitation. 

\subsection{Generalization to other ropes}
We tested if the learned model is successfully able to manipulate new ropes by qualitatively evaluating performance on a white jute rope that was significantly stiffer than the red rope used in the training and a black cotton rope that was significantly thinner and less stiff than the red rope used in the training. We found that the robot was successfully able to configure these ropes into relatively simpler ``L" and ``S" shapes. Even though the model was trained using interaction data from a single rope, it generalizes to other ropes. This shows that instead of learning features specific to the particular rope used for data collection, our model learns features that generalize to other ropes. 

One possible reason that the robot is unsuccessful at manipulating the white rope into more curvy ``W" and knot configurations is that the rope is too stiff to bent into curves necessary for forming the ``W" and the knot. With the black rope the robot was able to successfully manipulate the rope until the penultimate step in tying a knot, but failed at completing the knot. We also ran experiments where we changed the background from green to white and found that our model was unsuccessful at rope manipulation. This result is not surprising because all our training data is collected using a single background and it is expected that with training data collected on diverse backgrounds our method would be able to generalize to novel backgrounds. The video demonstrations of these experiments are available at the \href{https://ropemanipulation.github.io/}{\texttt{project website.}}  

\section{Discussion and Future Work}

In this paper, we presented a learning-based method for flexible manipulation of deformable objects such as ropes. Our method is based around self-supervision: we train a predictive model of rope behavior using data of rope interaction collected autonomously by the robot. This model predicts, given the current and target image of a rope, which action the robot can execute to put the rope into the target configuration. We combine our method with human demonstrations by allowing a human user to supply step-by-step images that show the rope undergoing a variety of manipulations. These demonstrations tell the system \emph{what} it should do with the rope at test time, while the learned model is used to determine \emph{how} to do it. Our experiments indicate that we can use our learned model to arrange the rope into a variety of different shapes using only high-level demonstration images.

One limitation of our approach is that, in its current form, it cannot learn to manipulate new objects exclusively by watching human demonstrations, since performing a manipulation task requires a model that can effectively predict the motion of the object, and this model is learned from the robot's own experience. In principle, this limitation could be overcome simply by collecting data from a large number of object manipulation scenarios, so as to learn a single model that generalizes effectively across objects. A more nuanced approach might involve correlating the behavior of objects in the human demonstrations with other previously experienced manipulations, so as to put them into correspondence and infer the behavior of an object for which prior experience is unavailable. In both cases, lifting this limitation is a promising direction for future work.

Although we demonstrate a variety of rope manipulation scenarios, our experiments are limited in scope, primarily due to limits on the amount of data that we can collect in reasonable time on a single robotic platform. For instance, most of our experiments use a single rope on a single background. Although we demonstrate that our model can successfully manipulate new ropes with significantly different material and texture properties into simple configurations, it fails to manipulate them into more complex configurations. 
If provided with substantially more robot-hours and a greater variety of ropes and environments, a similar model could in principle learn a more generalizable notion of rope manipulation. State-of-the-art results in image recognition and segmentation indicate that deep convolutional neural networks have the capacity to generalize to a wide range of scenes and objects when provided with enough data \cite{huh2016makes}, which suggests that the same kind of generalization could in principle be obtained by predictive models of deformable objects with enough data.

\section{Supplementary Materials and Videos}
Supplementary materials and videos of the robot's performance can be found at \href{https://ropemanipulation.github.io/}{\texttt{https://ropemanipulation.github.io/}}.

\section{Acknowledgements}

This work was supported in part by NSF CAREER award (\#IIS-1351028), NSF SMA-1514512, ONR MURI N00014-14-1-0671, and ONR YIP.

\addtolength{\textheight}{-9cm}

{\small
\bibliographystyle{IEEEtran}
\bibliography{rope}
}

\end{document}